\def\year{2021}\relax
\documentclass[letterpaper]{article} % DO NOT CHANGE THIS
\usepackage{aaai21}  % DO NOT CHANGE THIS
\usepackage{times}  % DO NOT CHANGE THIS
\usepackage{helvet} % DO NOT CHANGE THIS
\usepackage{courier}  % DO NOT CHANGE THIS
\usepackage[hyphens]{url}  % DO NOT CHANGE THIS
\usepackage{graphicx} % DO NOT CHANGE THIS
\usepackage{natbib}  % DO NOT CHANGE THIS AND DO NOT ADD ANY OPTIONS TO IT
\usepackage{caption} % DO NOT CHANGE THIS AND DO NOT ADD ANY OPTIONS TO IT
\usepackage{multirow}
\usepackage{float}
\usepackage{amsmath,amssymb} 
\usepackage{booktabs,tabu}
\usepackage[switch]{lineno}

\usepackage{color}

\title{SnapMix: Semantically Proportional Mixing for Augmenting Fine-grained Data}

\author {
        Shaoli Huang,\textsuperscript{\rm 1}
        Xinchao Wang, \textsuperscript{\rm 2}
        Dacheng Tao \textsuperscript{\rm 1} \\
}
\affiliations {
    \textsuperscript{\rm 1}  The University of Sydney \\
    \textsuperscript{\rm 2} Stevens Institute of Technology \\
    shaoli.huang@sydney.edu.au, xinchao.wang@stevens.edu , dacheng.tao@sydney.edu.au
}

\begin{document}
%\linenumbers
\maketitle

\begin{abstract}
Data mixing augmentation has proved effective in training deep models.
Recent methods mix labels mainly based on the mixture proportion of image pixels. 
As the main discriminative information of a fine-grained image usually resides in subtle regions, 
methods along this line are prone to heavy label noise in fine-grained recognition.
We propose in this paper a novel scheme, termed as
\textbf{S}ema\textbf{n}tic\textbf{a}lly \textbf{P}roportional \textbf{M}ixing (SnapMix),
which exploits class activation map (CAM) to lessen the label noise in augmenting fine-grained data. SnapMix generates the target label for a mixed image by estimating its intrinsic semantic composition, and allows for asymmetric mixing operations 
and ensures semantic correspondence between synthetic images and target labels. 
Experiments show that our method consistently outperforms existing mixed-based approaches  on various datasets and under different network depths.
Furthermore, by incorporating the mid-level features, the proposed SnapMix achieves top-level performance, demonstrating its potential to serve as 
a solid baseline for fine-grained recognition. Our code is available at https://github.com/Shaoli-Huang/SnapMix.git.
\end{abstract}

\section{Introduction}

\noindent Despite the remarkable success of deep neural networks,  
its overfitting problem persists, particularly in encountering 
limited training data. Data Augmentation methods can alleviate 
this by effectively exploiting existing data. Among them,  
mixing-based 
methods~\cite{tokozume2018between,inoue2018data,zhang2017mixup,yun2019cutmix} have
recently gained increasing attention.  
These approaches generate new data by blending images and fusing their labels according to the statistics of mixed pixels. 
For instance, Mixup~\cite{zhang2017mixup} combines images linearly and mix their targets using the same combination coefficients. CutMix~\cite{yun2019cutmix},
on the other hand,
cuts out one image area, pastes it on another image, and mix their labels according to the area proportion.  
By extending the training distribution, mixing-based techniques reduce memorizing data and improve model generalization. 

However, their superiority decreases with the 
increasing risk of label noise in augmenting fine-grained data. 
In fine-grained object recognition, 
discriminative information mainly lies in some small regions of images. 
Mixing labels based on mixture pixel-based statistics such as
area size,
therefore, tends to introduce severe label noise in this task. 
In the example  of Fig.~1, CutMix
cuts out a small region covering critical information about the label,
in this case a red shoulder and yellow wing bar of the red-winged blackbird.
The remaining part of the image, as a result, are left with only
much less informative image evidences, yet still 
take up a high coefficient due to its large area size.
This indicates that 
mixing labels based on area proportion 
is not able to effectively
reflect intrinsic semantic composition
of the combined image, 
thereby  deteriorating the data augmentation effectiveness
and confusing the model training.

In this paper, we propose 
a novel Semantically Proportional Mixing (SnapMix) 
strategy to address this issue.
SnapMix exploits a class activation map (CAM)~\cite{zhou2016learning}  to estimate
the label composition of the mixed-images. 
Specifically, by normalizing the CAM of each image
to sum to 1, we first obtain its Semantic Percent Map (SPM) to quantify the relatedness percentage between each pixel and the label,
and then compute
the semantic ratio of any image region 
by summing values in the corresponding area of the SPM.
For an image composed of multiple areas from multiple images, 
we can estimate its semantic composition through the semantic ratios 
corresponding to these regions.  
Compared to methods based on  statistics of mixture pixels, 
our label mixing strategy incorporates neural activations as prior knowledge to 
ensure the semantic correspondence between the synthetic images 
and the generated supervision signals.

Moreover,  
existing techniques rely on
\emph{symmetrically} blending image regions, 
meaning that the selected areas to be mixed are restricted to be complementary,
and hence limit the
diversity of augmented data.
By contrast, the proposed approach enables
\emph{asymmetric}
cut-and-paste operations,
allowing us to incorporate various factors such as
deformation and scale
into the data augmentation
to boost the data diversity. The current label-mixing strategies are designed based on the complementary principle. Thus they are not suitable for the \emph{asymmetric} operation that selects non-complementary regions to mix.

\begin{figure*}[t]
  \begin{center}
    \includegraphics[width=0.9\linewidth]{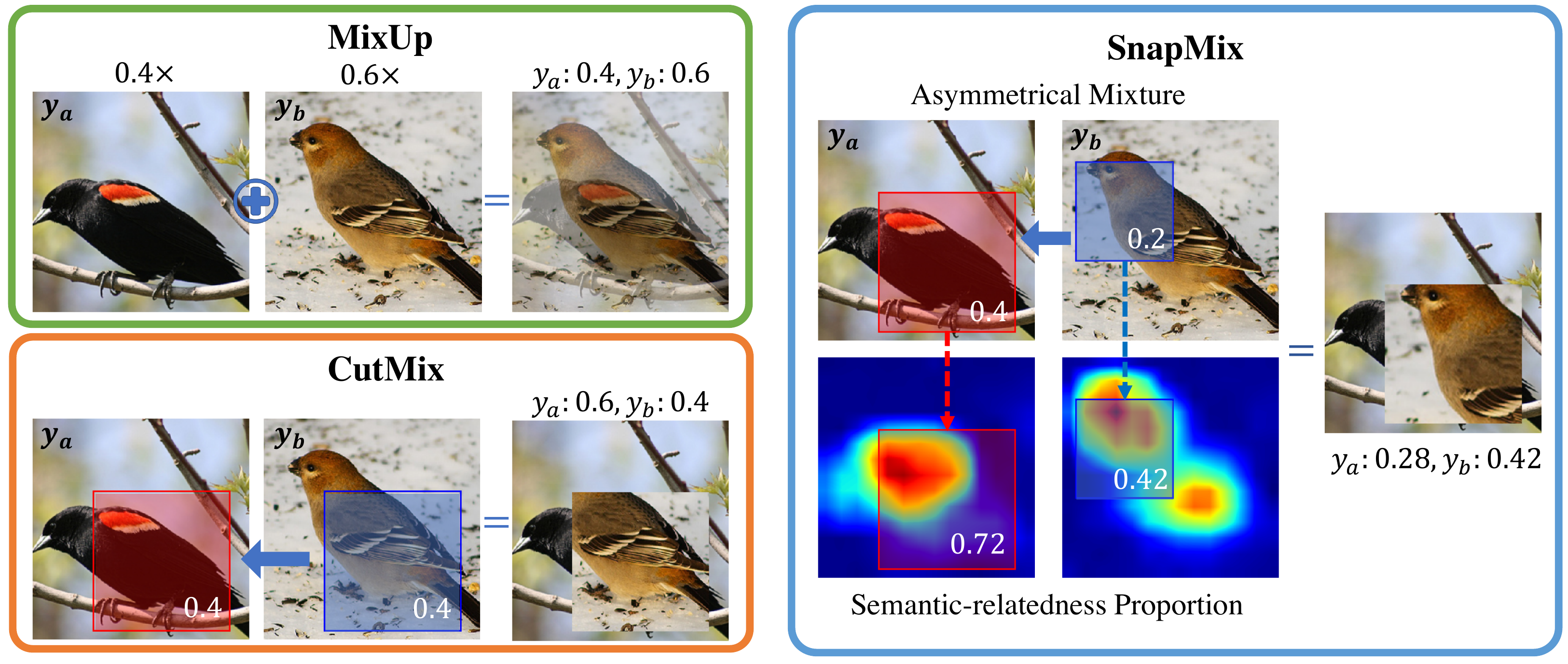}
  \end{center}
  \caption{Comparison of Mixup, CutMix, and SnapMix. The figure gives an example where SnapMix's generated label is visually more consistent with the mixed image's semantic structure comparing to CutMix and Mixup.}
%Technically, CutMix and Mixup mix images symmetrically and combines the labels based on the statistics of mixed pixels. In comparison, our method asymmetrically fuses images and estimates the semantic composition to combine labels.}
  \label{fig:title}
\end{figure*}

To validate the proposed approach, we adopt various network architectures (Resnet-18,34,50,101~\cite{he2016deep}) as baseline models and 
compare our method with existing data augmentation approaches 
on three fine-grained datasets. 
Results indicate that prior methods
lead to unstable performances, sometimes even harmful,
when using shallow network architecture.
This can be in part explained by the fact 
shallow neural networks are not able
to well tackle  label noises,
which significantly degrade data augmentation effectiveness.
The proposed method, on the other hand,
consistently outperforms compared methods
on various datasets and with different network depths. 
Furthermore, we show that even a simple model can achieve comparable state-of-the-art performance when applying our proposed data augmentation. 
This indicates that our method can well serve as a  solid
baseline for advancing fine-grained recognition.

\section{Related Works}

\subsection{Fine-Grained Classification} 
Fine-grained recognition has been an active research area in recent years. This task is more challenging than general 
image classification~\cite{LiuTPAMI16,Wang11TIP,Yang18TNNLS,Yang_2020_CVPR,yang2020factorizable}, as the critical information to distinguish categories usually lies in subtle object parts. Part-based methods thereby, are extensively explored to address the problem. Early works  \cite{huang2016part,zhang2014part, xiao2015application,lin2015deep,xu2015augmenting,xu2016friend} mainly rely on strongly supervised learning to localize object part for subsequent feature learning. Due to part annotations are expensive to acquire, the later methods \cite{zhang2016picking, Zheng2017,sun2018multi, zheng2019looking} attempts to find discriminative part regions in a weakly supervised manner. For example, Zhang et al. \cite{zhang2016picking} first picks distinctive filters and then use them to learn part detectors through an iteratively alternating strategy. MA-CNN \cite{Zheng2017}  obtains part regions by clustering feature maps of intermediate convolutional layers, MAMC \cite{sun2018multi}. In recent years, fine-grained approaches have developed in the direction of enforcing the neural networks to acquire rich information \cite{yang2018learning, ding2019selective, chen2019destruction, zhang2019learning}. For instance, Zhang et al.,\cite{zhang2019learning}  progressively crop out discriminative regions to generate diversified data sets for training network experts. Chen et al., \cite{chen2019destruction} destruct the images and then learn a region alignment network to restore the original spatial layout of local regions. These works implicitly integrate data augmentation practices into their methodologies, which relate to our proposed method mostly.  

However, our proposed method SnapMix differs from them in two aspects. First, SnapMix is a pure data augmentation based technique that does not require an extra computational process in the testing stage. Besides, our approach builds on recent advances from the data mixing strategy. In contrast, those methods are mainly based on conventional data augmentation strategy, which typically processes a single image and retains the original label.

\subsection{Data augmentation} 
Recent advances \cite{takahashi2019data,zhong2017random,devries2017improved,tokozume2018between,inoue2018data,zhang2017mixup,yun2019cutmix}  in data augmentation can be divided into two groups: region-erasing based and data mixing.  The former~\cite{zhong2017random,devries2017improved} erases partial region of images in training, aiming to encourage the neural networks to find more discriminative regions. The typical method is CutOut that augments data by cutting a rectangle region out of an image. The other line of methods is data mixing based \cite{tokozume2018between,inoue2018data,zhang2017mixup} that have recently gained increasing attention in the field of image classification. Compared with region-erasing augmentation, these methods generate new data by combing multiple images and fusing their labels accordingly. Among those works, Zhang et al., \cite{zhang2017mixup}  first proposed mixing data to extend the training distribution. Their proposed method termed as MixUp, generated images by linearly combining images and fusing their labels using the same coefficients. MixUp showed its superiorities in handling corrupted targets and improving model performance.  Summers and Dineen \cite{summers2019improved} further improved Mixup by introducing a more generalized form of data mixing that considered non-linear mixing operations. In very recent work, Yun et al. proposed CutMix \cite{yun2019cutmix} that produces a new image by cutting out one image patch and pasting to another image. Similar to Mixup, the labels are also mixed but proportionally to the area of the patches. By taking advantage of both types of methods, CutMix showed impressive performance in classification tasks and weakly-supervised localization tasks. 

Our proposed method falls into the second category. However, it differs significantly from the previous techniques in the way of mixing labels.  Current mixing-data based approaches combine labels mainly depending on the statistic of mixture pixels, such as the ratio of pixel number or intensity values. In comparison, our method estimates the semantic structure of a synthetic image by exploiting class activation maps. This new characteristic allows our approach to augment fine-grained data without introducing severe label noise. Another slight difference is that SnapMix blends images using asymmetric patches,  resulting in better data randomness and diversity than those using symmetric regions.

\begin{figure*}[t]
  \begin{center}
    \includegraphics[width=1.0\linewidth]{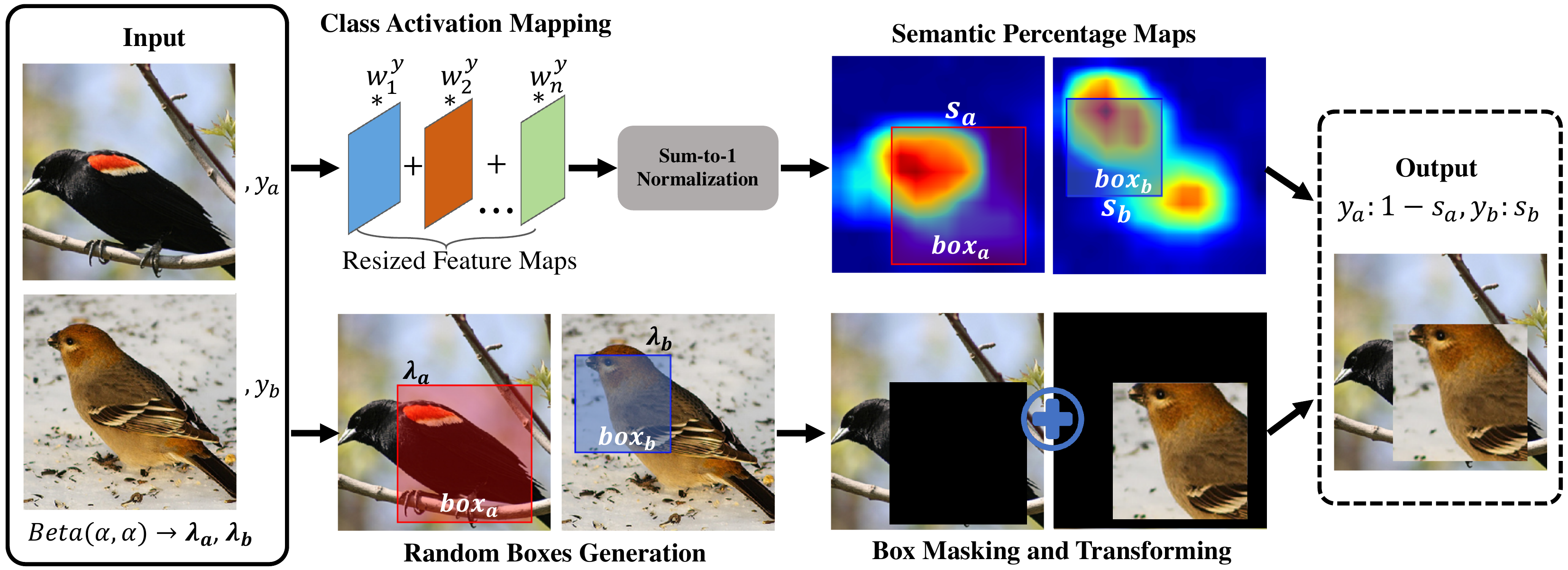}
  \end{center}
  \caption{An overview of proposed method.}
  \label{fig:overview}
\end{figure*}

\section{Semantically Proportional Mixing}

Data augmentation has become an indispensable step for training deep neural networks. The standard augmentation methods mainly apply a composition of image preprocessing techniques on an input image,  such as flipping, rotations, color jittering, and random cropping.  Recent works demonstrated the great potential of mixing-based techniques for training deep models. Unlike standard practice, these methods generate new data by combining images and also mixing the corresponding labels. In the following, we first provide some notations used in this paper. We then briefly introduce two representative mix-based approaches Mixup and CutMix. Next, we describe in detail our proposed method SnapMix.

\subsection{Notations}
We use the following notations throughout this paper. The original training data set  $\{(I_i,y_i) | i\in [0,1,...,N-1]\}$, where $I_i \in R^{3\times W\times H}$ and $y_i$ refer to an input image and the label respectively. Given a data pair $((I_a,y_a),(I_b,y_b))$ and hyperpatemer $\alpha$, mixing-based methods first draw a random value $\lambda$ from a beta distribution $Beta(\alpha,\alpha)$. Then they generate a new image $\tilde{I}$ and two label weights $\rho_a$ and $\rho_b$ according to $\lambda$. Here, $\rho_a$ and $\rho_b$ are corresponding to the label $y_a$ and $y_b$ respectively.  %We further define weight matrices $P_i$ and $P_j \in R^{W\times H}$  that describe how $x_i$ and $x_j$'s pixels are used in mixing. Thus, we can express the image mixing operations in a general form , that is
\iffalse
\begin{equation}
    \tilde{x} = x_i \odot P_i + x_j \odot P_j,
\end{equation}
where $\odot$ denotes element-wise multiplication. Here, different mix-based methods construct $P_i$ and $P_j$ in a different way.
\fi
\subsection{Mixup and CutMix}

Recent mixing-based methods essentially stem from two representative techniques Mixup and cutMix. 

\textit{MixUp} mixes images and combines labels using linear combination, which is expressed as

\begin{equation}
\begin{aligned}
    &\tilde{I} = \lambda \times I_a  + (1-\lambda) \times I_b,\\
    &\rho_a = \lambda, \rho_b = 1 - \lambda,
 \end{aligned} 
\end{equation}

%Thus, $P_i$ and $P_j$ can be expressed as 
%\begin{equation}
%    P_i = \lambda \times J,  P_j = (1-\lambda)\times J.
%    \label{eq:mixup}
%\end{equation}

\textit{CutMix} adopts cut-and-paste operation for mixing images and mixes the labels according to the area ratio. That is

\begin{equation}
\begin{aligned}
    &\tilde{I} = (1-M_{\lambda}) \odot I_a  +  M_{\lambda}\odot I_b,\\
    &\rho_a = 1-\lambda, \rho_b = \lambda,
 \end{aligned} 
\end{equation}
where $\odot$ denotes element-wise multiplication and $M_{\lambda}\in R^{W\times H}$ is a binary mask of a random box region whose area ratio to the image is $\lambda$.

On one hand, these two methods mainly differ in the way they mix images.  Mixup mixes image by linear combination and therefore improves the neural networks' robustness to adversarial examples. By integrating Mixup and regional dropout strategy, the cut-and-paste regime of Cutmix naturally inherits this advantage but also enhances models' capabilities of object localization.  On the other hand, they share two similarities: 1) mixing labels by using the statistic of mixture pixels, and 2) performing image mixing in symmetric locations.

\iffalse
Now, we can give a general form of calculating they label weights for both Mixup and Cutmix, which is defined as
\begin{equation}
    %w_i =\frac{sum(P_i)}{sum(P_i)+sum(P_j)}, w_j =\frac{sum(P_j)}{sum(P_i)+sum(P_j)}
    w_i =\frac{sum(P_i)}{sum(J)}, w_j =\frac{sum(P_j)}{sum(J)}
    \label{eq:weight}
\end{equation}
By substituting Eq.\ref{eq:mixup} and Eq.\ref{eq:cutmix} into Eq.\ref{eq:weight}  respectively, our definitions obtain consistent results both with Mixup and CutMix. Also, the formulation shows that both Mixup and Cutmix compute the label weights only based on the statistics of mixture pixels.
\fi

\subsection{SnapMix for Fine-grained Recognition}

In fine-grained recognition, the category difference usually resides in subtle object parts, making part localization ability plays an important role. Therefore, the cut-and-paste mechanism is more favorable in augmenting fine-grained data. However, mixing labels by the region area ratio is unreliable and will increase the risk of label noise, particularly in combining image at asymmetric locations. Motivated by work that used class activation maps (CAMs) to describe the class-specific discriminative regions, we propose to exploit CAMs to estimate the semantic composition of a mixed image. Fig.~\ref{fig:overview} shows an overview of our proposed method SnapMix.  Our proposed method differs existing methods in two folds: 1) fusing labels based on semantic composition estimation, 2) mixing images asymmetrically. Given an input pair of data, we first extract their semantic percentage maps used to compute the semantic percentage of any image area. We then mix the images by cut-and-paste at asymmetrical locations. Finally, we calculate each mixture component's semantic proportion as guidance to fuse the one-hot labels. In the following, we further describe in detail our method in terms of image mixing and label generation.

\textbf{Mixing images.} As discussed previously, current existing methods blend images at symmetric locations, limiting the diversity of synthetic images. Our approach removes this constrain to increase the randomness of data augmentation further. Specifically, instead of using a single random location, we crop an area at a random location in one image and transform and paste it to another random place in another image. Such mixing operation is expressed as

\begin{equation}
    %\tilde{I} = T_{\theta}(M^a_{\lambda} \odot I_a)  + (1-M^b_{\lambda}) \odot I_b,
    \tilde{I} = (1-M_{\lambda^a}) \odot I_a  + T_{\theta}(M_{\lambda^b} \odot I_b),
    \label{eq:snapmix}
\end{equation}
 where $M_{\lambda^a}$ and $M_{\lambda^b}$ are two binary masks containing random box regions with the area ratios $\lambda^a$ and $\lambda^b$, and $T_{\theta}$ is a function that transforms the cutout region of $I_b$ to match the box region of $I_a$.

\textbf{Label generation.}
To estimate the semantic composition of a mixed image, we need to measure each original image pixel's semantic relatedness to the corresponding label. One alternative to do this can resort to class activation map,  as it is proved useful to interpret how a region correlates with a semantic class. Thus, we first employ the attention method\cite{zhou2016learning} to compute the class activation maps of input images. For a given image $I_i$, we denote $F(I_i) \in \mathbb{R}^{d \times h \times w}$ the output of the last convolutional layer,  $F_l(I_i)$ the $l^{th}$ feature map of $F(I_i)$, and $w_{y_i} \in \mathbb{R}^{d}$ the classifier weight corresponding to class $y_i$. Then we can obtain $I_i$'s class activation map $CAM(I_i)$ by 

\begin{equation}
    CAM(I_i) = \Phi(\sum_{l=0}^{d}w_{y_i}^lF_l(I_i)),
\end{equation}
where $\Phi(\cdot)$ denotes a operation that upsamples a feature map to match dimensions with input image size. Here, we ignore the bias term for simplicity.
We can now obtain a Semantic Percent Map (SPM) by normalizing the CAM to sum to one. Here, we define SPM as a semantic information measure map to quantify the relatedness percentage between a pixel and the label.  We compute the SPM of an image $S(I_i)$ by

\begin{equation}
    S(I_i) = \frac{CAM(I_i)}{sum(CAM(I_i))},
\end{equation}

Finally, for an image $\title{I}$ produced using Eq.\ref{eq:snapmix}, we compute the corresponding label weights $\rho_a$ and $\rho_b$ as

\begin{equation}
\begin{aligned}
    %w_i &=\frac{sum(P_i\odot M(x_i))}{sum(P_i\odot M(x_i))+sum(P_j\odot M(x_j))}, \\
    %w_j &=\frac{sum(P_j\odot M(x_j))}{sum(P_i\odot M(x_i))+sum(P_j\odot M(x_j))}
    %w_i =\frac{sum(P_i\odot M(x_i))}{sum(M(x_i))},w_j =\frac{sum(P_j\odot M(x_j))}{sum(M(x_j))}
    %w_i =\frac{sum(P_i\odot M(x_i))}{sum(M(x_i))},w_j =\frac{sum(P_j\odot M(x_j))}{sum(M(x_j))}
    %w_i =sum(P_i\odot M(x_i)),w_j =sum(P_j\odot M(x_j))
    %&\rho_a =sum(M^a_{\lambda}\odot S(I_a)),\\
    %&\rho_b =1-sum(M^b_{\lambda}\odot S(I_b)).
    &\rho_a =1-sum(M_{\lambda^a}\odot S(I_a)),\\
    &\rho_b =sum(M_{\lambda^b}\odot S(I_b)).
    \end{aligned}
\end{equation}
By doing so,  the generated supervision information for a mixed image can better reflect its intrinsic semantic composition. Therefore, in fine-grained recognition, despite the image's discriminative information is extremely uneven in spatial distribution, our method prevent introducing heavy noise in the augmented data.

It is also worth noting that the two components of a mixed image generally do not complement each other in terms of semantic proportion.  A case of this would be when a cutout is a background patch and pasted over the object area of another image, and then the synthesized image would not contain any foreground object. Therefore, unlike CutMix, our method does not restrict the label coefficients ($\rho_a$ and  $\rho_b$)  to sum up to 1.

\iffalse
\begin{equation}
    M(x_i) = \frac{C(x_i)}{sum(C(x_i))}
\end{equation}

\begin{equation}
    M(x_i) = \frac{C(x_i)}{sum(C(x_i)+sum(C(x_j)))}
\end{equation}
\fi

  \begin{table*}
    % \scriptsize
    % \small
    % \tiny
    % \scriptsize
        \caption{Performance comparison(Mean Acc.$\%$) of methods using backbone networks \textit{Resnet-18} and \textit{Resnet-34} on fine-grained datasets. Each method's improvement over the baseline is shown in the brackets.  }
        \begin{center}
            \begin{tabular}{lcc|cc|cc}
            % \begin{tabu} to 0.8\textwidth {@{}X[c]XXXX@{}}
            \toprule
            %{\multicolumn{4}{c}{Accuracy(\%)} \\
            %\cmidrule{2-4} 
            %& \multicolumn{3}{c}{CUB} & \multicolumn{3}{c}{Cars} & \multicolumn{3}{c}{Aircraft} \\
            & \multicolumn{2}{c}{CUB} & \multicolumn{2}{c}{Cars} & \multicolumn{2}{c}{Aircraft} \\
            \midrule
            %&\multicolumn{3}{c}{ResNet-34} &\multicolumn{3}{c}{ResNet-50} &\multicolumn{3}{c}{ResNet-101}\\
            &Res18 & Res34  &Res18 & Res34 &Res18 & Res34  \\
            Baseline & 82.35 & 84.98 &91.15  & 92.02 & 87.80 & 89.92  \\
            \hline
            CutOut & 80.54 (-1.81) & 83.36 (-1.62)  & 91.83(+0.68) & 92.84 (+0.82) & 88.58 (+0.78) & 89.90 (-0.02) \\
             MixUp              & 83.17 (+0.82)  & 85.22 (+0.24) & 91.57 (+0.42) & 93.28 (+1.26) & 89.82 (+2.02) & 91.02 (+1.1) \\
          %         \hspace{6mm} +SC          & 85.09 & 86.16 & 0& 0 & 0 & 0& 0 & 0 & 0\\
           % \hline

          %  \hline
             CutMix            & 80.16 (-2.19)& 85.69 (+0.71) &  92.65 (+1.50) & 93.61 (+1.59)& 89.44 (+1.64) & 91.26 (+1.34) \\
             \hline
             SnapMix            & \textbf{84.29 (+1.94)}  & \textbf{87.06 (+2.08)} &\textbf{93.12(+1.97)} &  \textbf{93.95 (+1.93)} &\textbf{90.17 (+2.37)}  & \textbf{92.36 (+2.44)}\\
        %     \hspace{6mm}+SC   & 86.31$^{\textbf{+0.62}}$ & 87.38$^{\textbf{+1.23}}$ & 0& 0 & 0 & 0& 0 & 92.30 & 0\\
            \bottomrule
            \end{tabular}
            % \end{tabu}
        \end{center}
        \label{table:augmentation}
        % \vspace{-10mm}
  \end{table*}
  
    \begin{table*}
    % \scriptsize
    % \small
    % \tiny
    % \scriptsize
        \caption{Performance comparison(Mean Acc.$\%$) of methods using backbone networks \textit{Resnet-50} and \textit{Resnet-101} on fine-grained datasets. Each method's improvement over the baseline is shown in the brackets.}
        \begin{center}
            \begin{tabular}{lcc|cc|cc}
            % \begin{tabu} to 0.8\textwidth {@{}X[c]XXXX@{}}
            \toprule
            & \multicolumn{2}{c}{CUB} & \multicolumn{2}{c}{Cars} & \multicolumn{2}{c}{Aircraft} \\
            \midrule
            & Res50 & Res101  & Res50 & Res101& Res50 & Res101 \\
            %\hline
            Baseline & 85.49 & 85.62 & 93.04 & 93.09 & 91.07 & 91.59 \\
            \hline
             CutOut              & 83.55 (-1.94) & 84.70 (-0.92) & 93.76 (+0.72) & 94.16 (+1.07) & 91.23 (+0.16) & 91.79 (+0.2)\\
             MixUp                & 86.23 (+0.74) & 87.72 (+2.1) & 93.96 (+0.92) & 94.22 (+1.13) & 92.24 (+1.17) & 92.89 (+1.3)\\
            CutMix             & 86.15 (+0.66) & 87.92 (+2.3) & 94.18 (+1.14) & 94.27 (+1.18) & 92.23 (+1.16) & 92.29 (+0.7)\\
             \hline
             SnapMix             & \textbf{87.75 (+2.26)}& \textbf{88.45 (+2.83)} & \textbf{94.30 (+1.21)} & \textbf{94.44 (+1.35)}& \textbf{92.80 (+1.73)} & \textbf{93.74 (+2.15)}\\
            \bottomrule
            \end{tabular}
            % \end{tabu}
        \end{center}
        \label{table:augmentation}
        % \vspace{-10mm}
  \end{table*}

\section{Experiments}

In this section, we extensively evaluated the performance of SnapMix on three fine-grained datasets. We evaluated our method using multiple network structures  (Resnet-18,34,50,101) as baselines. We compared the performance of our approach and related data augmentation methods on each network architecture. Further, we tested our method using a strong baseline that integrated mid-level features and compared the results with those of the current state-of-the-art methods of fine-grained recognition.

\iffalse
\begin{table}
    % \scriptsize
    % \small
    % \tiny
    % \scriptsize
        \caption{Hyperameters used for CutOut,MixUp,and CutMix}
        \begin{center}
            \begin{tabular}{lcc}
            \toprule
            Method & $p$ & $\alpha$ \\
            \midrule
            CutOut & 0.5 & -  \\
            MixUp & 0.5 & 1.0  \\
            CutMix & 1.0 & 3.0 \\
             \bottomrule

            \end{tabular}
            % \end{tabu}
        \end{center}
        \label{table:augmentation}
        % \vspace{-10mm}
  \end{table}
\fi
\subsection{Datasets}
We conduct experiments on three standard fine-grained  datasets, which are CUB-200-2011 \cite{wah2011caltech}, Stanford-Cars \cite{krause20133d}, and FGVC-Aircraft \cite{maji2013fine}. For each dataset, We first resized images to $512 \times 512$  and cropped them with size  $448 \times 448$. In the rest of the paper, we used the short names CUB, Cars, and Aircraft to simplify the notation.

\iffalse
\textit{CUB} dataset is one of the most challenging datasets in fine-grained recognition. The dataset contains 200 bird species, each of which has roughly 60 images. The training set has 5994 samples, and the test set has 5794 samples.

\textit{ Cars} dataset consists of 16,185 images containing 196 car models. For this dataset, 8144 images are for training and the rest for testing.

\textit{Aircraft} dataset includes 102 aircraft models and 10,200 images. Each airplane model has 100 images. This dataset uses 6667 images for training and the rest for testing.
\fi
\subsection{Experiment Setup}

\textbf{Backbone networks and baselines.}
To extensively compare our method with other approaches,  we used four network backbones as baselines in performance comparison. Here, if not specified, we refer baseline as a neural network model that was pre-trained on Imagenet dataset and fine-tuned on a target dataset. The used network structures include Resnet-18,34,50 and 101. Here, we adapted their implementation from the TorchVision package to our experiments.

  \begin{table*}[t]
   % \scriptsize
   % \small
   % \tiny
   % \scriptsize
   
   \begin{center}
       % \begin{tabular}{|l|l|p{2.0cm}p{2.0cm}p{2.0cm}|}
       \begin{tabu} to 0.9\textwidth {@{}p{6.5cm}p{2.0cm}XXX@{}}
       \toprule
       {\multirow{2}{*}{Method}} & {\multirow{2}{*}{Backbone}} & \multicolumn{3}{c}{Accuracy(\%)} \\
       \cmidrule{3-5}
       & & CUB & Cars & Aircraft\\
   
       \midrule
       % VGG-19 & VGG-19 & 77.8 & 84.9 & 84.8\\

      % STN \cite{jaderberg2015spatial}  & $5 \times \text{Inception}$ & 84.1 & - & -\\ 
      % Compact B-CNN \cite{gao2016compact} & $1 \times \text{VGG-16}$ & 84.0 & -  & - \\
     %   MA-CNN \cite{Zheng2017} & $ 5 \times \text{VGG-19}$ & 86.5 & 91.5 & 89.9\\
       RA-CNN \cite{Fu2017} & $3 \times \text{VGG-19}$ & 85.3 & 92.5 & 88.2 \\
       RAM \cite{li2017dynamic} & $3 \times \text{Res-50}$ & 86.0 & - & \\ 
       %Low-rank B-CNN \cite{kong2017low} &  $1 \times \text{VGG-16}$ & 84.2 & 90.9 & 87.3 \\
       %Kernel-Activation \cite{cai2017higher} &  $1 \times \text{VGG-16}$ & 85.3 & 91.7 & 88.3 \\
       Kernel-Pooling \cite{cui2017kernel} & $1 \times \text{VGG-16}$ & 86.2 & 92.4 & 86.9 \\
       NTS-Net \cite{yang2018learning} & $5 \times \text{Res-50}$ & 87.5 & 93.9 & 91.4 \\
       DFL-CNN \cite{wang2018learning} & $1 \times \text{VGG-16}$ & 86.7 & 93.8 & 92.0 \\
       MAMC \cite{sun2018multi} & $1 \times \text{Res-101}$ & 86.5 & 93.0 & - \\
       DFL-CNN \cite{wang2018learning} & $1 \times \text{Res-50}$ & 87.4 & 93.1 & 91.7 \\
       DCL \cite{chen2019destruction} & $1 \times \text{Res-50}$ & 87.8 & {94.5} & {93.0} \\
       TASN \cite{zheng2019looking} & $1 \times \text{Res-50}$ & 87.9 & 93.8 & - \\
       %\midrule
       %Mixup \cite{zhang2017mixup} & $1 \times \text{Resnet-50}$ & 87.80 & 94.14 & 92.35\\
       %Cutout \cite{devries2017improved} & $1 \times \text{Resnet-50}$ & 86.74 & 94.74 & 92.23\\
       %CutMix \cite{yun2019cutmix} & $1 \times \text{Resnet-50}$ & 87.88 & 94.64 & 92.77\\
        S3N \cite{ding2019selective} & $3 \times \text{Res-50}$ & {88.5} & {94.7} & 92.8 \\
       MGN-CNN \cite{zhang2019learning}  & $3 \times \text{Res-50}$ & {88.5} & 93.9 & - \\
       MGN-CNN \cite{zhang2019learning}  & $3 \times \text{Res-101}$ & \textbf{89.4} & 93.6 & - \\
       \midrule
       baseline & $1 \times \text{Res-50}$ & 85.49 (85.85) & 93.04 (93.17) & 91.07 (91.30)\\
    baseline$^{\dagger}$ & $1 \times \text{Res-50}$ & 87.13  &  93.80&91.68  \\
        baseline & $1 \times \text{Res-101}$ & 85.62 (86.02) & 93.09 (93.28) & 91.59 (92.11)\\
        baseline$^{\dagger}$ & $1 \times \text{Res-101}$ &87.81& 93.94 &91.85  \\
        \midrule
        baseline  \hspace{2mm}+ \textbf{SnapMix} & $1 \times \text{Res-50}$ & 87.75 (88.01) & 94.30 (94.59) & 92.80 (93.16)\\
        baseline$^{\dagger}$  \hspace{0.6mm}+ \textbf{SnapMix} & $1 \times \text{Res-50}$ & 88.70 (88.97) &\textbf{95.00}(95.16) & 93.24(93.49)\\
         baseline  \hspace{2mm}+ \textbf{SnapMix} & $1 \times \text{Res-101}$ & 88.45 (88.73) & 94.44 (94.60) &  93.74 (94.03)\\
        baseline$^{\dagger}$  \hspace{0.6mm}+ \textbf{SnapMix} & $1 \times \text{Res-101}$ &89.32 (89.58)  & 94.84 (94.96) &\textbf{94.05} (94.24) \\
       \bottomrule
       % \end{tabular}
       \end{tabu}
   \end{center}
   \caption{The accuracy ($\%$)  comparison with state-of-the-art methods on CUB, Cars, and Aircraft. For the baselines and our approach, we reported their average accuracy of the final ten epochs and showed their best accuracy in the brackets.}
   \label{table:results}
   % \vspace{-4mm}
   \end{table*}

We also used a strong baseline that incorporates mid-level features in performance evaluation. Here, we termed it \textbf{baseline$^{\dagger}$}. This baseline was used in recent works\cite{wang2018learning,zhang2019learning} to push the performance limits of fine-grained recognition. Compared with the standard baseline that contains a single classification branch, Baseline$^{\dagger}$ adds another mid-level classification branch on top of the intermediate layers.  In our experiments, we followed the implementation from \cite{zhang2019learning}. Specifically, the mid-level branch included a \textit{Conv1$\times 1$}, \textit{Max Pooling}, and a \textit{Linear Classifier layer} and was placed after $4^{th}$ block of ResNet.  We blocked the gradients passing the mid-level branch to backbone networks in training. In testing, we fused the predictions from two classification branches.

\noindent\textbf{Data augmentation methods.}
We compared our method with three representative data augmentation methods namely CutOut \cite{devries2017improved}, MixUp \cite{zhang2017mixup}, and CutMix \cite{yun2019cutmix}. Since these works did not officially report results on fine-grained datasets,  we implemented these methods based on the released codes and run experiments on fine-grained datasets. We first tested different hyperparameters for each method and then selected the optimal one for all network structures.  We set the probability of performing augmentation $0.5$  for CutOut and MixUp and $1.0$ for CutMix. We used the $\alpha$ values of $1.0$ and $3.0$ for MixUp and CutMix, respectively.

\noindent\textbf{Training details.}
We used stochastic gradient descent (SGD) with momentum 0.9, base learning rate 0.001 for the pre-trained weights, and 0.01 for new parameters. We trained our model for 200 epochs and decayed the learning rate by factor 0.1 every 80 epochs.

%We implemented our method based on PyTorch \cite{paszke2017automatic} and trained it on Nvidia V100 GPU. In training, we initialized backbone networks with Imagenet pre-trained models from the TorchVision package. We used stochastic gradient descent (SGD) with momentum 0.9, base learning rate 0.001 for the pre-trained weights, and 0.01 for new parameters. We trained our model for 200 epochs and decayed the learning rate by factor 0.1 every 80 epochs. We employed some common practices of fine-grained datasets in training. We first resized images to $512 \times 512$  and cropped them with size  $448 \times 448$. We adopted standard data augmentations, including random cropping and randomly horizontal flipping. We used the same training details for both our method and other data augmentation approaches.

\subsection{Performance evaluation}
In this section, we presented the results of our method and performance comparisons with existing approaches. We first made comparisons between SnapMix and other data augmentation methods.  Further, we tested our approach using the two baselines and compared the results with those of the current state-of-the-art methods. We used top-1 accuracy as the performance measure and provided both the best accuracy and average accuracy (the mean result of the final $10$ epochs) of our proposed method.
%For most of the existing results reported on fine-grained datasets, it is unclear whether they were obtained by selecting the best epoch. Therefore, we provided both the best accuracy and average accuracy (the mean result of the final $10$ epochs).

\noindent\textbf{Comparison with data augmentation methods.}
We listed the results of performance comparisons in Table. 1-2. Here, Table 1-2 shows each method's average accuracy and improvement over the baseline. First, we can observe that our proposed method SnapMix consistently outperforms its counterparts on three datasets. We can further find that existing methods mostly yield limited, even negative improvement on the CUB dataset. This might mainly because the CUB dataset exhibits more subtle category differences, making those methods increase the risk of noise labels. Besides, the effectiveness of these methods is relatively sensitive to the network depth. For example,  both Mixup and CutMix achieve significant improvements on the CUB dataset only using the deeper networks Resnet-101, and CutMix even suffers a performance drop when using Resnet-18.  We hypothesis the reason is that the deeper models have better capacities in handling label noise. In comparison,  SnapMix significantly improves the baseline regardless of network depth.

\noindent\textbf{Comparison with state-of-the-art methods.}

In this section, we compared the performance of SnapMix and other state-of-the-art techniques of fine-grained recognition.  In Table. \ref{table:results},  first, we can observe that the baseline$^{\dagger}$ achieved higher accuracy than the baseline on three datasets, and the performance gain on the CUB dataset is the most significant. This result indicates mid-level features can effectively complement the capacity of the global-level features in fine-grained recognition. It is also worth mentioning that some top-performing works, such as DFL-CNN \cite{wang2018learning} and MGN-CNN \cite{zhang2019learning} also embedded the baseline+ into their methods. 

Secondly, SnapMix enhances both baselines to obtain comparable performance even to some latest approaches with intricate designs and high inference time.  S3N \cite{ding2019selective}  and MGN-CNN \cite{zhang2019learning} are two of the state-of-the-art methods. S3N adopted a selective sparse sampling strategy to construct multiple features. MGN-CNN exploit attention mechanisms to construct different inputs for multiple expert networks. Both methods require a similar data processing pipeline with the training stage and the need for multiple feed-forward passes of the backbone network in the testing stage.
\iffalse
%\textbf{Comparing with state-of-the-art methods}
\begin{figure}[t]
  \begin{center}
    \includegraphics[width=0.9\linewidth]{LaTeX/curve.pdf}
  \end{center}
  \caption{Test accuracy curve comparison of different augmentation methods.}
  \label{fig:curve}
\end{figure}
\fi

In contrast, using a standard baseline with a single Resnet-101 backbone, SnapMix, without bells and whistles in the testing stage,  achieves the accuracy of $88.45\%$, $94.44\%$, and $93.74\%$ on CUB, Cars, and Aircraft respectively, which outperforms most of the existing techniques. Even using the baseline$^{\dagger}$ (a more powerful baseline), SnapMix still demonstrates its promise and effectiveness in performance improvement, pushing the accuracy to the next level.  For example, SnapMix achieves $89.32\%$ accuracy (close to the result of MGN-CNN $89.4\%$)   on the CUB dataset and exhibits superior performance than all the comparing techniques on both Cars and Aircraft dataset.

\subsection{Analysis}

%In this section, we first analyzed the impact of the hyperparameter on the performance of SnapMix.  Then we examined the effectiveness of each component of SnapMix. Finally, we compare the testing accuracy curves to understand the regularization ability of different data augmentation methods. We conducted all the following ablation experiments based on the typical dataset CUB and backbone network ResNet-50.

\noindent\textbf{Training from scratch.} Tab.~\ref{table:scratch} shows that our approach is also effective without using ImageNet pre-trained weights. In this experiment, we used the 'switch' probability (the probability of applying the mixing augmentation) of 0.5 for each mixing method. This allows the networks to learn from both clean and mixed data, preventing the mixed data from excessively affecting the model's initial learning stage. Therefore, despite SnapMix may introduce noise labels in the early training stage, it would not hinder the network from learning a good CAM in the subsequent stage. This is because the network tends to first learn from easy samples (clean data ) other than difficult samples (mixed data with label noise) \cite{arpit2017closer}. With the continuous learning of the network and the improvement of CAM quality, the more reasonable the label estimated by SnapMix will be to enhance subsequent model learning.

  \begin{table}
    % \scriptsize
    % \small
    % \tiny
    % \scriptsize
        \caption{Performance comparison of training from scratch on the CUB dataset (Acc.$\%$).}
        \begin{center}
            \begin{tabular}{ccccc}
            \toprule
            &Baseline&CutMix&MixUp&SnapMix  \\
             \midrule
             Res-18 & 64.98 & 60.03 & 67.63 & \textbf{70.31} \\
             Res-50 & 66.92 & 65.28 & \textbf{72.39} & 72.17 \\
             \bottomrule

            \end{tabular}
            % \end{tabu}
        \end{center}
        \label{table:scratch}
        % \vspace{-10mm}
  \end{table}

\noindent\textbf{Effectiveness of using other network backbones.} 
 We evaluated the performance of our method with two other network backbones including IceptionV3 \cite{szegedy2016rethinking} and DenseNet121 \cite{huang2017densely}. As shown in Tab~\ref{table:othernet},  our method surpasses both CutMix and MixUp approaches and improves the baseline by a large margin. This result demonstrates SnapMix's consistent effectiveness when applied to various CNN architecture.
  \begin{table}
    % \scriptsize
    % \small
    % \tiny
    % \scriptsize
        \caption{Performance comparison of using other network backbones on the CUB dataset (Acc.$\%$).}
        \begin{center}
            \begin{tabular}{ccccc}
            \toprule
            &Baseline&Cutmix&Mixup&Snapmix  \\
             \midrule
             InceptionV3 & 82.22 & 84.31 & 83.83 & \textbf{85.54} \\
             DenseNet121 & 84.23 & 86.11 & 86.65 & \textbf{87.42} \\
             \bottomrule

            \end{tabular}
            % \end{tabu}
        \end{center}
        \label{table:othernet}
        % \vspace{-10mm}
  \end{table}

\noindent\textbf{Influence of hyperparameters.}
The hyperparameter $\alpha$ of snapMix decides a beta distribution that is used to generate a random patch in mixing. To investigate its impact on the performance, we tested seven values of $\alpha$.
 Table.\ref{table:influence} showed that the accuracy increased slightly with the increase of $\alpha$ value and peaked at the number of $5$, which suggests the importance of using the medium-size boxes to mix images on this dataset. Besides, the accuracy of setting different $\alpha$ values inconsiderably fluctuates around the mean value of $87.37\%$, indicating that snapMix is not very sensitive to the $\alpha$ value.

\begin{figure}[t]
  \begin{center}
    \includegraphics[width=1\linewidth]{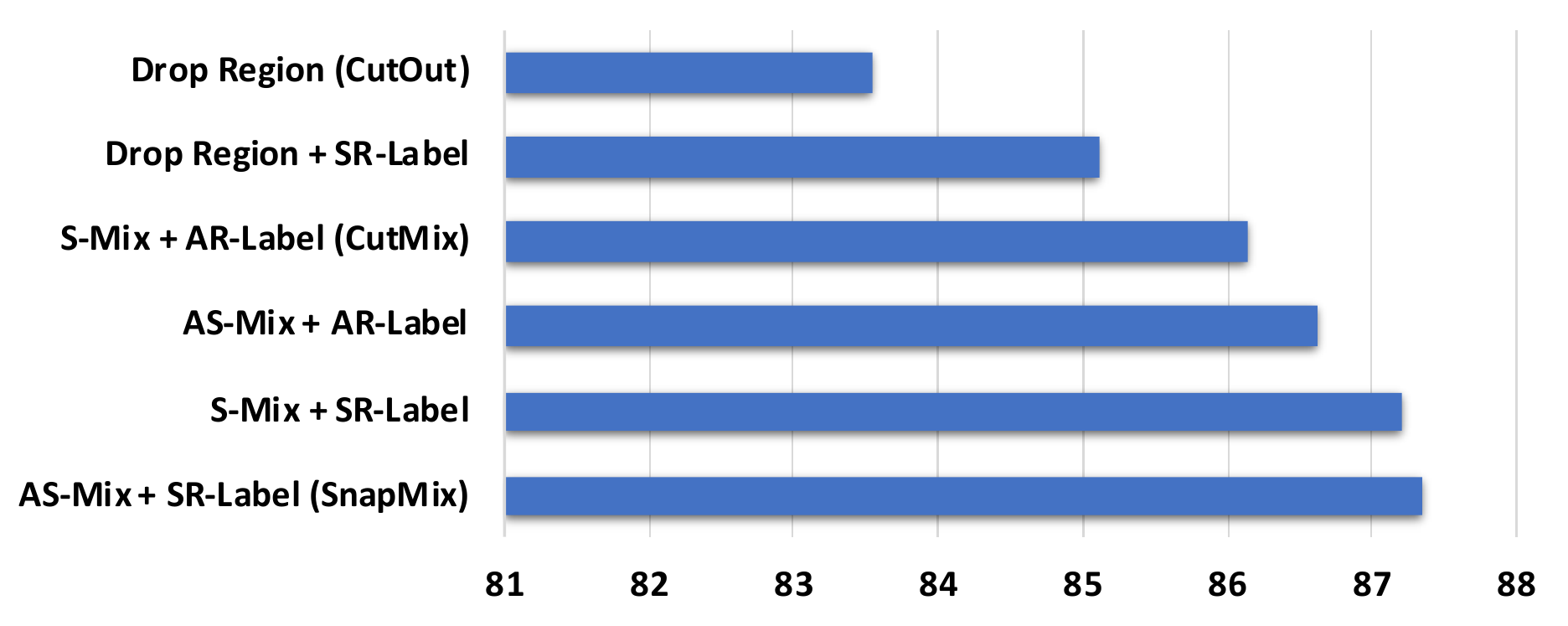}
  \end{center}
  \caption{Accuracy comparison of six different combination techniques ($\%$). Here, \textbf{S-Mixing}, \textbf{AS-Mix}, \textbf{AR-label}, and \textbf{SR-label}  are short for symmetric mixing, asymmetric mixing, area ratio label, and semantic ratio label respectively.}
  \label{fig:acc}
\end{figure}

\noindent\textbf{Effectiveness of each component of SnapMix.} We performed experiments using combinations of different image mixing operations and label mixing strategies. As shown in Fig.~\ref{fig:acc}, the asymmetric mixing provides a slight improvement over the symmetric mixing, and the label mixing strategy of SnapMix is the primary contributor to the performance gain. More importantly, the Semantic-Ratio consistently shows improvement in using three image mixing operations. 

%SnapMix mainly contains two technical components: Asymmetric image mixing and Semantically proportional label mixing. To understand their effectiveness individually, we performed experiments using combinations of different image mixing operations and label mixing strategies.  Here, we used three image mixing functions from CutMix, SnapMix, and CutOut, namely symmetric mixing (S-Mix, asymmetric mixing (AS-Mix), and drop regions. We tested two label combining strategies: Area-Ratio based (CutMix) and Semantic-Ratio based (Snapmix).  

%Therefore, we had six combinations in this experiment and showed the result in Fig.~\ref{fig:acc}.  First, we can see that asymmetric mixing provides a slight improvement over the symmetric mixing, and the label mixing strategy of SnapMix is the primary contributor to the performance gain. More importantly, the Semantic-Ratio consistently shows improvement in using three image mixing operations. For example, it improved Drop Region from the accuracy of $83.55\%$ to $85.12\%$, S-Mix from $86.15\%$ to $87.12\%$, and AS-Mix from $86.62\%$ to $87.75\%$.

  \begin{table}
    % \scriptsize
    % \small
    % \tiny
    % \scriptsize
        \caption{Influence of hyperparameters Acc.($\%$)}
        \begin{center}
            \begin{tabular}{ccccccc}
            \toprule
            $\alpha$= 0.2&0.5 &1.0 &3.0 & 5.0 & 7&8  \\
             \midrule
             87.22 & 87.23& 87.25& 87.30&87.75 &87.30&87.54  \\

             \bottomrule

            \end{tabular}
            % \end{tabu}
        \end{center}
        \label{table:influence}
        % \vspace{-10mm}
  \end{table}
  
  \begin{figure}[!]
  \begin{center}
    \includegraphics[width=0.9\linewidth]{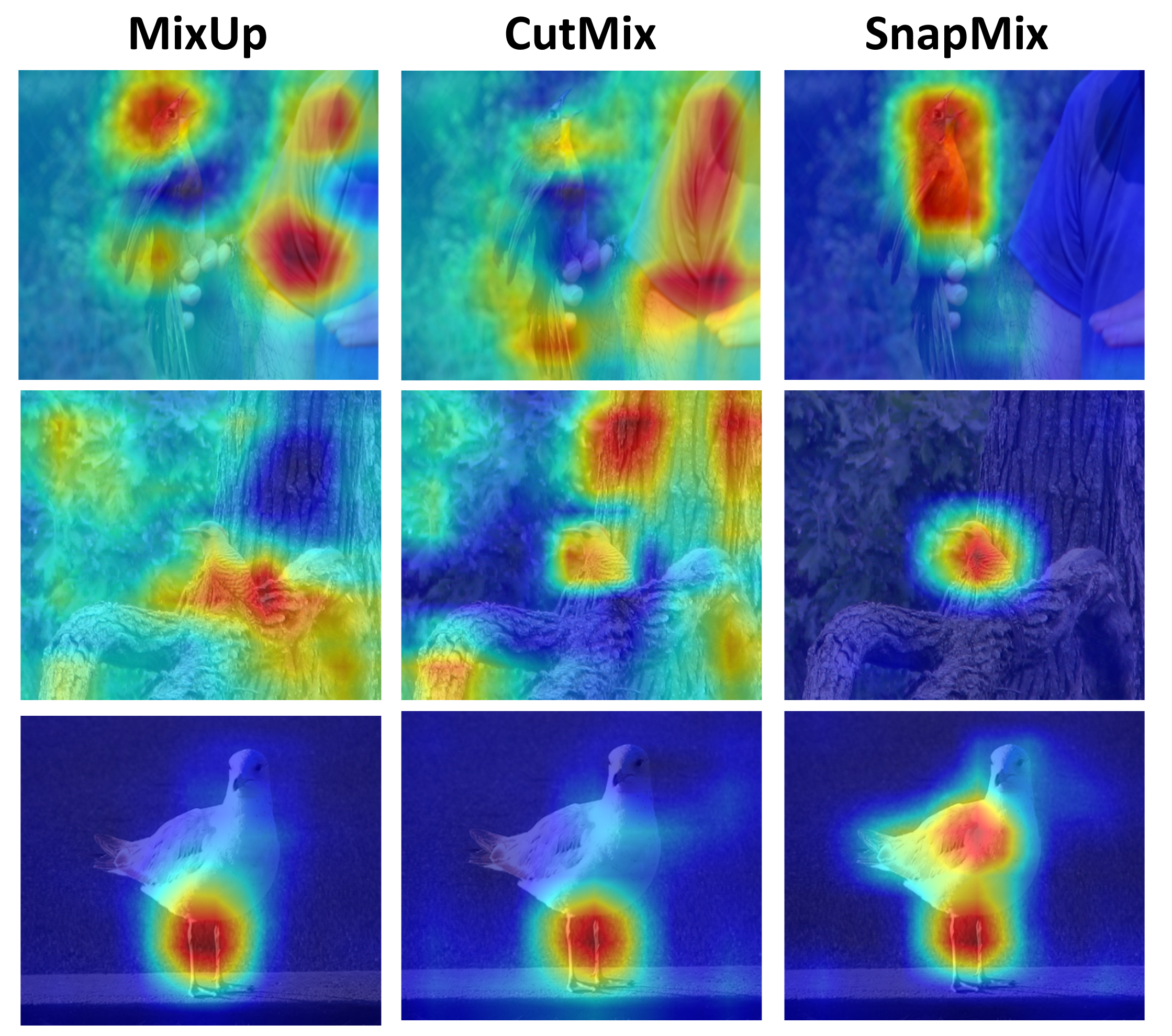}
  \end{center}
  \caption{CAM visualization of different augmentation methods.}
  \label{fig:vis}
  \vspace{-1em}
\end{figure}
  
  \noindent\textbf{Visualization.}
Fig.~\ref{fig:vis} shows CAMs of some examples correctly predicted by SnapMix but misclassified by MixUp and CutMix. We can observe the attention of MixUp and CutMix are distracted by some background patterns, which might be a reason for the misprediction. By comparison, the network attention of SnapMix tends to lie in object regions. These results imply mixing labels by pixel statistics may cause the neural networks more sensitive to background visual patterns, while our proposed method avoids this issue.

\section{Conclusions}
In this paper, we present a new method SnapMix for augmenting fine-grained data. SnapMix generates new training data with more reasonable supervision signals by considering the semantic correspondence. Our experiments showed the importance of estimating semantic composition for a synthetic image.  Our proposed method might also benefit other tasks (e.g.,  indoor scene recognition or person re-identification), where a small image region contains significant discriminative information. The proposed label mixing strategy is mainly applicable to cut-and-paste mixing. Further work might explore how better to estimate the semantic structure of a linearly combined image.
\section{Acknowledgements}
This work was supported by the Australian Laureate Fellowship project FL170100117, DP-180103424, IH-180100002,
and Stevens Institute of Technology Startup Funding.
 \bibliography{egbib}
\bibliographystyle{aaai21}
\end{document}